\documentclass[sigconf,nonacm]{acmart}
\renewcommand\footnotetextcopyrightpermission[1]{} %remove copyright
\usepackage{multirow}
\usepackage{pifont}
\newcommand{\cmark}{\ding{51}}%
\newcommand{\xmark}{\ding{55}}%
\usepackage{colortbl}

\begin{document}

\title{CFSR: Geometry-Conditioned Shadow Removal via Physical Disentanglement}

\author{Pan Wang}
\authornote{Equal contribution.}
\authornote{Corresponding author.}
\affiliation{%
  \institution{University of Science and Technology of China}
  \city{Hefei}
  \country{China}
}
\email{diogenescask@gmail.com}

\author{Yihao Hu}
\authornotemark[1]
\affiliation{%
  \institution{Ant Group}
  \city{Hangzhou}
  \country{China}
}

\author{Xiujin Liu}
\affiliation{%
  \institution{University of Michigan - Ann Arbor}
  \city{Ann Arbor}
  \country{USA}
}

\author{Hang Wang}
\affiliation{%
  \institution{Sun Yat-sen University}
  \city{Zhuhai}
  \country{China}
}

\begin{abstract}
Traditional shadow removal networks often treat image restoration as an unconstrained mapping, lacking the physical interpretability required to balance localized texture recovery with global illumination consistency. To address this, we propose CFSR, a multi-modal prior-driven framework that reframes shadow removal as a physics-constrained restoration process. By seamlessly integrating 3D geometric cues with large-scale foundation model semantics, CFSR effectively bridges the 2D-3D domain gap. Specifically, we first map observations into a custom HVI color space to suppress shadow-induced noise and robustly fuse RGB data with estimated depth priors. At its core, our Geometric \& Semantic Dual Explicit Guided Attention mechanism utilizes DINO features and 3D surface normals to directly modulate the attention affinity matrix, structurally enforcing physical lighting constraints. To recover severely degraded regions, we inject holistic priors via a frozen CLIP encoder. Finally, our Frequency Collaborative Reconstruction Module (FCRM) achieves an optimal synthesis by decoupling the decoding process. Conditioned on geometric priors, FCRM seamlessly harmonizes the reconstruction of sharp high-frequency occlusion boundaries with the restoration of low-frequency global illumination. Extensive experiments demonstrate that CFSR achieves state-of-the-art performance across multiple challenging benchmarks.
\end{abstract}


\ccsdesc[500]{Computing methodologies~Computer vision tasks; Image representations; Appearance and texture representations}

\keywords{3D geometric priors, semantic priors, feature disentanglement.}



\maketitle

\section{Introduction}
\label{sec:intro}

\begin{figure}[t]
    \centering
    \includegraphics[width=\linewidth]{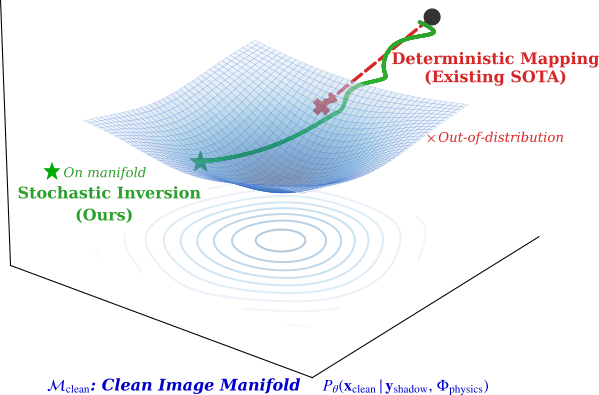}
    \caption{
        \textbf{Conceptual comparison of shadow removal paradigms.} 
        Unlike existing deterministic mapping approaches (red dashed line) that suffer from out-of-distribution artifacts, our method reframes shadow removal as a \textbf{physics-constrained degradation inversion}  (green solid line), ensuring the restored image strictly adheres to the latent clean image manifold $\mathcal{M}_{\text{clean}}$.
    }
    \label{fig:teaser_concept}
    \vspace{-0.2cm}
\end{figure}

Shadow removal is a fundamental yet challenging task in computer vision, serving as a crucial pre-processing step for high-level downstream applications such as object detection~\cite{tran2024low,hu2025sde, wang2025multi}, tracking~\cite{bai2026drwkv}, and autonomous driving~\cite{wang2023ultra}. While recent deep learning-based approaches have achieved remarkable progress, the majority of existing networks formulate this task as a deterministic 'image-to-image' mapping problem. This 'black-box' paradigm inherently lacks the physical interpretability required to balance spatial detail recovery with global illumination consistency. Operating primarily within the 2D RGB domain, these methods remain oblivious to the underlying 3D physical geometry~\cite{liu2025gnc,liu2021shadow,guo2023shadowformer,le2019shadow}. Consequently, they struggle with the ill-posed nature of the task, frequently failing to distinguish between intrinsic dark textures and actual shadows cast by complex geometries.

\begin{figure}[t]
    \centering
    \includegraphics[width=\linewidth]{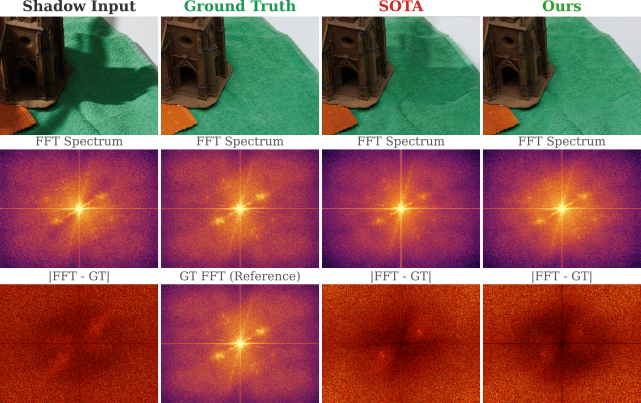}
    \caption{
        \textbf{Visual and spectral comparison against SOTA~\cite{lin2025densesr}.} 
        From top to bottom: restored images, corresponding 2D FFT magnitude spectra, and absolute spectral discrepancy ($|\text{FFT} - \text{GT}|$, brighter red indicates larger errors). Our physics-constrained inversion paradigm preserves structural integrity and yields the minimal spectral error.
    }
    \label{fig:fft_analysis_detailed}
    \vspace{-0.2cm}
\end{figure}

Fundamentally, shadow generation is not merely a localized darkening effect, but rather a geometry-dependent physical phenomenon driven by intricate light absorption and scattering processes~\cite{1542031}. Existing architectures lack the physics-aware framework necessary to invert these degradations, particularly in the frequency domain. Due to the absence of physics-constrained inversion logic, they often compromise global lighting consistency for local details, resulting in conspicuous visual artifacts~\cite{9601181}. 
To break this bottleneck, we argue that shadow removal must be reframed from a deterministic blind pixel regression to a \textbf{physics-constrained degradation inversion problem}. As illustrated in Fig.~\ref{fig:teaser_concept}, rather than adopting a deterministic mapping that easily falls into out-of-distribution states, our ultimate goal is to mathematically reverse the complex light-matter interaction. This paradigm ensures that the restored image adheres strictly to the physical manifold of the latent clean scene ($\mathcal{M}_{\text{clean}}$). 
Such a physics-guided framework effectively mitigates the global-local trade-off bottleneck. As demonstrated by the spectral analysis in Fig.~\ref{fig:fft_analysis_detailed}, compared to existing methods, our degradation inversion strictly reverses the degradation, thereby preserving the structural integrity in the spatial domain and significantly reducing the spectral discrepancy ($|\text{FFT} - \text{GT}|$) in the frequency domain.

To this end, we propose \textbf{CFSR}, a multi-modal prior-driven shadow removal framework that bridges the 2D-3D domain gap by transforming conventional image-to-image mapping into a \textbf{physics-constrained restoration process}. By seamlessly integrating estimated 3D geometric priors with large-scale foundation model semantics, CFSR effectively overcomes the inherent ill-posed limitations of single-modal reconstruction. Specifically, we first employ a pre-trained monocular depth estimator to derive explicit 3D geometric cues. To robustly fuse these cross-domain modalities, we introduce a custom \textbf{HVI color space} equipped with a learnable \textbf{Intensity Collapse mechanism}. This design dynamically suppresses severe noise in deeply shadowed regions while constructing a reliable multi-modal feature base.

At the core of our architecture lies the \textbf{Geometric \& Semantic Dual Explicit Guided Attention (GSGA)}. Diverging from conventional implicit feature learning, this module utilizes semantic correlation maps from a pre-trained DINO~\cite{oquab2023dinov2} and planar correlation maps derived from 3D surface normals to explicitly modulate the attention affinity matrix~\cite{DepthAnythingV2}. This structurally enforces physical lighting constraints during representation aggregation. Furthermore, to combat severe information loss in dense shadows, we leverage a frozen CLIP visual encoder at the deep bottleneck~\cite{radford2021learning}. By injecting globally robust semantic priors via cross-modal attention, the network is guided to hallucinate and recover lost textures with holistic scene awareness. Finally, the decoding phase is driven by the \textbf{Frequency Collaborative Reconstruction Module (FCRM)}. Instead of monolithic regression, FCRM elegantly decouples high-frequency texture refinement from low-frequency illumination smoothing under continuous 3D geometric conditioning, achieving a mathematically rigorous balance between high-fidelity structural preservation and global color consistency.

In summary, the main contributions of this work are three-fold:
\begin{itemize}
    \item \textbf{Physics-Aware Formulation \& Representation:} We fundamentally advance shadow removal from an unconstrained 2D mapping task to a physics-conditioned restoration framework. By introducing a custom HVI space with an Intensity Collapse mechanism, we enable the dynamic suppression of shadow-induced noise and the seamless fusion of degraded RGB observations with explicit 3D geometric priors.
    
    \item \textbf{Geometric \& Semantic Guided Extraction:} We propose the Dual Explicit Guided Attention mechanism. By aligning dense 2D semantics (DINO) with spatial 3D surface normals, we explicitly inject physical lighting and structural constraints directly into the attention affinity matrix, effectively bridging the 2D-3D modality gap.
    
    \item \textbf{Geometry-Conditioned Collaborative Decoding:} We design the Frequency Collaborative Reconstruction Module (FCRM) to disentangle the synthesis process into specialized high- and low-frequency branches. Coupled with a CLIP-guided global semantic injection, this module explicitly incorporates 3D planar constraints to resolve geometry-dependent illumination degradation, ensuring an optimal balance between photorealistic texture recovery and physical consistency.
\end{itemize}

\section{Related Work}
\label{sec:related}

\subsection{Image Shadow Removal}
Shadow removal has been extensively studied due to its critical role in enhancing scene understanding and downstream vision tasks. Early traditional methods primarily relied on physical illumination models and hand-crafted priors, such as gradient domain manipulation ~\cite{liu2008texture}, illumination transfer~\cite{6241432}, and region-based color matching~\cite{guo2011single}. While theoretically sound, these methods often fail in complex real-world scenes due to the fragility of hand-crafted features. 
With the advent of deep learning, convolutional neural networks (CNNs) and Generative Adversarial Networks (GANs) have dominated this field. Representative works~\cite{wang2018stacked,Hu_2019_ICCV,jiang2021focal} formulate shadow removal as a deterministic image-to-image translation problem, employing various attention mechanisms and multi-scale feature fusions to map shadowed inputs to shadow-free outputs. Recently, Transformer-based architectures~\cite{guo2023shadowformer,liu2024regional,xiao2024homoformer} have been introduced to capture long-range dependencies, further improving restoration quality. 
\textbf{Limitation \& Our Difference:} Despite their success, these methods predominantly operate as 'black-box' deterministic mappings within the 2D RGB domain. They lack the physical interpretability to model the complex light-matter interactions, often leading to a compromise between local texture recovery and global illumination consistency. In contrast, our CFSR reframes this task as a physics-constrained degradation inversion problem, breaking the deterministic 2D mapping paradigm.

\subsection{Physical Priors and Geometry-Aware Vision}
Shadows fundamentally originate from the physical occlusion of light transport by 3D scene geometry~\cite{liang2025diffusion,liu2025gnc,yang2023sire,einy2022physics,zhang2021no}. Consequently, treating shadow removal as a purely 2D pixel-level mapping inevitably suffers from inherent ambiguities, such as the classic challenge of distinguishing dark surface albedo from shaded regions. To alleviate such ill-posed problems, recent advances in intrinsic image decomposition and relighting have increasingly leveraged large-scale physical priors, including depth maps, surface normals, and semantic features~\cite{DepthAnythingV2,oquab2023dinov2,kirillov2023segment,li2025lirm}. \textbf{Limitation \& Our Difference:} While recent attempts have begun introducing auxiliary modalities into shadow removal, they predominantly rely on simple implicit feature concatenation~\cite{guo2023boundary,hu2025shadowhack,chen2025prompt}. They fail to mathematically constrain the network using the underlying 3D structural and lighting principles. Our work bridges this 2D-3D domain gap explicitly and structurally. First, rather than fusing modalities in the ambiguous RGB space, we project the observation into a physics-aware HVI color space, employing a novel Intensity Collapse mechanism to robustly fuse color representations with estimated 3D geometric priors (depth and normals). More importantly, instead of treating these priors as mere input channels, our Geometric \& Semantic Dual Explicit Guided Attention (GSGA) fundamentally rewires the representation learning. By explicitly modulating the attention affinity matrix with 3D planar correlations and semantic boundaries, GSGA directly injects strict physical laws into the feature aggregation process, ensuring illumination consistency across complex 3D surfaces.

\subsection{Foundation Models and Semantic Priors}
The emergence of large-scale foundation models, such as CLIP~\cite{radford2021learning} and DINO~\cite{oquab2023dinov2}, has revolutionized visual representation learning. Trained on massive datasets, these models encapsulate robust, high-level semantic priors that have been increasingly transferred to various low-level vision tasks, including image restoration~\cite{11353407}, super-resolution~\cite{11369964}, and inpainting~\cite{11370297}. Unlike purely pixel-driven networks, foundation models offer illumination-invariant feature manifolds, providing crucial holistic context when local observations are severely degraded.
\noindent \textbf{Limitation \& Our Difference:} While foundation models provide rich semantics, directly applying them to shadow removal—a task intrinsically coupled with 3D physics—often results in hallucinated artifacts that violate local scene geometry. Furthermore, standard feature aggregation mechanisms fail to balance the competing demands of detail synthesis and global illumination. To address this, CFSR innovatively aligns foundation model priors with explicit physical constraints. Specifically, we utilize frozen DINO and CLIP encoders to inject semantic priors, compensating for severe information loss in dense shadows. More importantly, rather than relying on unconstrained feature blending, we introduce the Frequency Collaborative Reconstruction Module (FCRM). By disentangling the representation space and explicitly conditioning the synthesis on 3D geometric structures, FCRM effectively harnesses these powerful semantic priors while preventing physically implausible hallucinations, achieving a rigorous balance between texture fidelity and physical consistency.

\section{Method}
\label{sec:method}

The primary goal of our proposed CFSR is to reframe shadow removal from a purely 2D image-to-image translation task into a physics-constrained restoration process. To achieve this, we design a multi-modal prior-driven framework that seamlessly integrates 3D geometric properties (depth and surface normals) and large-scale foundation model semantics (DINO and CLIP). 

\subsection{Overall Architecture}
\label{subsec:overall_architecture}

\begin{figure*}[t]
    \centering
    \captionsetup{skip=-1pt}
    \includegraphics[width=\textwidth]{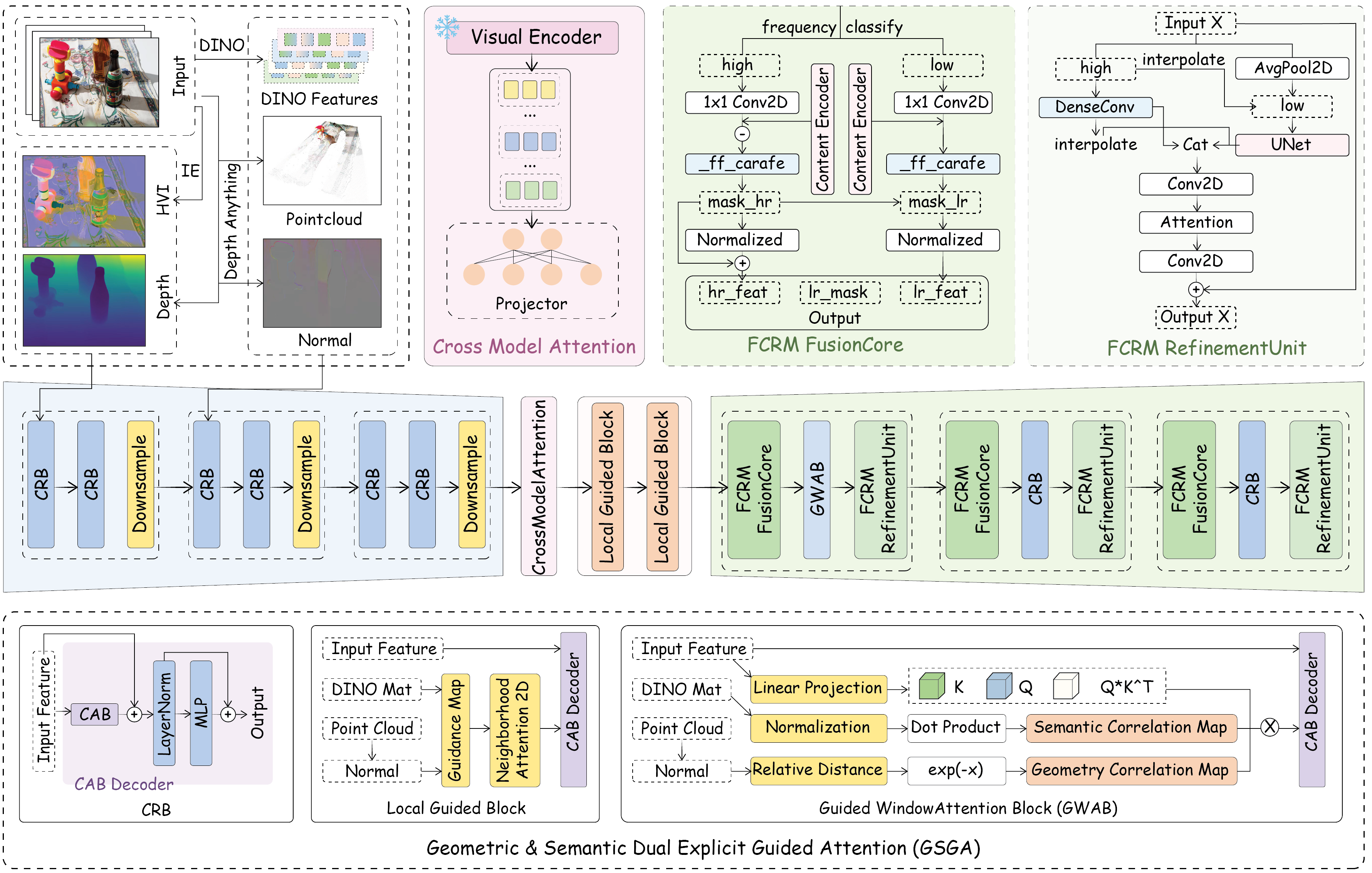}
    \caption{
    \textbf{The overall architecture of the proposed CFSR.} 
    Given a shadowed image, CFSR first extracts 3D geometric (Point Cloud, Normals) and 2D semantic (DINO) priors. A frozen visual encoder (CLIP) injects robust global semantics via Cross-Modal Attention at the bottleneck. During decoding, the Geometric \& Semantic Dual Explicit Guided Attention (GSGA) strictly constrains feature aggregation using 3D spatial and semantic correlations. Finally, the Frequency Collaborative Reconstruction Module (FCRM) disentangles and refines high-/low-frequency components to achieve physics-consistent restoration.
    }
    \label{fig:architecture}
\end{figure*}

The overall architecture of CFSR is illustrated in Fig.~\ref{fig:architecture}. Given a shadowed input image $\mathbf{I}_{in} \in \mathbb{R}^{3 \times H \times W}$, the pipeline consists of four main stages:

\noindent\textbf{1) 3D Prior Estimation \& Semantic Extraction:} 
To bridge the 2D-3D domain gap, we employ a pre-trained monocular depth estimator to extract 3D geometric priors from $\mathbf{I}_{in}$, yielding a depth point cloud $\mathbf{P} \in \mathbb{R}^{3 \times H \times W}$ and a surface normal map $\mathbf{N} \in \mathbb{R}^{3 \times H \times W}$. Simultaneously, $\mathbf{I}_{in}$ is fed into a frozen pre-trained DINOv2 network to extract robust region-level semantic features $\mathbf{F}_{dino}$. 

\noindent\textbf{2) Multi-Modal Feature Initialization:} 
To decouple complex illumination from chrominance, we project $\mathbf{I}_{in}$ into a custom HVI color space with a learnable \textit{Intensity Collapse} mechanism to dynamically suppress shadow noise, constructing a robust multi-modal feature base $\mathbf{F}_{0}$.

\noindent\textbf{3) Physics-Guided Feature Extraction:} 
The feature is processed by a U-Net backbone equipped with the \textit{Dual Explicit Guided Attention (GSGA)}, which explicitly enforces physical constraints. Global semantic priors from CLIP are injected at the bottleneck.

\noindent\textbf{4) Frequency Collaborative Reconstruction Module (FCRM):} 
Finally, FCRM disentangles the deep features into specialized high- and low-frequency branches, ensuring a mathematically rigorous balance between high-fidelity texture recovery and physical consistency.

\subsection{3D Prior Estimation \& Semantic Extraction}
\label{subsec:prior_estimation}

Our framework assumes that shadow degradation is heavily conditioned by the underlying 3D scene geometry and semantic context. Therefore, before entering the main restoration network, we explicitly construct a physical geometry field and a high-dimensional semantic manifold from the 2D shadowed input $\mathbf{I}_{in} \in \mathbb{R}^{3 \times H \times W}$.

\noindent \textbf{Physical Geometry Field.} 
We first estimate the pseudo-depth map $\mathbf{D} \in \mathbb{R}^{1 \times H \times W}$ using a pre-trained monocular estimator. Since precise camera intrinsics are unavailable for standard shadow datasets (e.g., ISTD, SRD), we employ a normalized pseudo-intrinsic matrix $\mathbf{K} \in \mathbb{R}^{3 \times 3}$ assuming a fixed field-of-view ($60^\circ$) and a centered principal point. This relative scaling is sufficient for extracting local planar variations. Given $\mathbf{K}$ and pixel coordinates $(u, v)$, the 3D spatial point $\mathbf{p}(u, v) \in \mathbb{R}^3$ is derived via inverse perspective mapping:
\begin{equation}
    \mathbf{p}(u, v) = \mathbf{D}(u, v) \cdot \mathbf{K}^{-1} [u, v, 1]^T
\end{equation}
This yields the dense 3D point map $\mathbf{P} \in \mathbb{R}^{3 \times H \times W}$. To explicitly model structural constraints for illumination, we compute the surface normal field $\mathbf{N} \in \mathbb{R}^{3 \times H \times W}$. For a local $5 \times 5$ neighborhood $\Omega$, the normal $\mathbf{n}$ is the eigenvector corresponding to the smallest eigenvalue of the local covariance matrix $\mathbf{C}$:
\begin{equation}
    \mathbf{C}(u,v) = \frac{1}{|\Omega|} \sum_{(i,j) \in \Omega} (\mathbf{p}(i,j) - \bar{\mathbf{p}})(\mathbf{p}(i,j) - \bar{\mathbf{p}})^T
\end{equation}
where $\bar{\mathbf{p}}$ is the centroid in $\Omega$. To mitigate scale ambiguities caused by pseudo-intrinsics, the extracted normals are strictly $L_2$-normalized. This provides robust, scale-invariant physical constraints for the subsequent representation learning.

\noindent \textbf{High-dimensional Semantic Manifold.} 
To inject illumination-invariant priors, $\mathbf{I}_{in}$ is upsampled via a bilinear operator $\mathcal{U}$ and mapped into a semantic manifold using a frozen DINOv2 encoder, $\Phi_{dino}$. To preserve dense spatial semantics, we aggregate representations from intermediate Transformer blocks:
\begin{equation}
    \mathbf{F}_{dino} = \text{Concat}_{l \in \mathcal{L}} \Big( \mathcal{P}_l \big( \Phi_{dino}^{(l)}( \mathcal{U}(\mathbf{I}_{in}) ) \big) \Big) \in \mathbb{R}^{C_{dino} \times H \times W}
\end{equation}
where $\Phi_{dino}^{(l)}$ denotes the output of the $l$-th intermediate block, and $\mathcal{L}$ is the set of selected layers. $\mathcal{P}_l$ acts as a spatial projection operator, performing bilinear interpolation for spatial alignment followed by a $1 \times 1$ convolution to reduce the channel dimension to $C_{dino}/|\mathcal{L}|$. Here, $\text{Concat}$ denotes channel-wise concatenation. 

Ultimately, the tuple $\mathcal{T}_{prior} = \{ \mathbf{P}, \mathbf{N}, \mathbf{F}_{dino} \}$ constitutes our multi-modal condition, providing explicit physical and semantic guidance for the subsequent restoration process.

\subsection{Multi-Modal Feature Initialization}
\label{subsec:feature_initialization}

Standard RGB color spaces intrinsically entangle illumination and chrominance, complicating the decoupling of shadow-induced degradation. To address this, we transform the input $\mathbf{I}_{in}$ into a physics-aware, polar-transformed representation. Specifically, we extract the structural illumination map (Value $V = \max(R,G,B)$) and saturation ($S$). To circumvent the angular discontinuity of standard hue ($H \in [0, 1]$), we project it into continuous polar components:
\begin{equation}
    H_{x} = S \cdot \cos(2\pi H), \quad H_{y} = S \cdot \sin(2\pi H)
\end{equation}

\noindent \textbf{Intensity Collapse Mechanism (ICM).}
While extracting the pure illumination channel ($V$) is effective, deeply shadowed regions (where $V \to 0$) inevitably suffer from severe signal-to-noise ratio (SNR) degradation. Direct feature extraction in these dark areas critically amplifies hidden sensor noise. Rather than employing complex spatial gating networks, we introduce an elegant, non-linear \textit{Intensity Collapse Mechanism} (ICM). We apply a learnable power-law transformation to squelch noise fluctuations dynamically:
\begin{equation}
    \hat{V} = \max(V, \epsilon)^{\kappa}, \quad \text{where} \;\; \kappa = \ln(1 + e^{k_{hvi}}) + \epsilon
\end{equation}
where $k_{hvi}$ modulates the learnable decay factor $\kappa$, and $\epsilon = 10^{-6}$. The optimized coefficient $\kappa > 1.0$ enforces a steep non-linear thresholding effect. This isolates and diminishes low-magnitude sensor noise inherent to dark regions, thereby decoupling authentic structural gradients from shadow-induced artifacts.

Subsequently, the noise-suppressed photometric tensor $\hat{\mathbf{I}}_{HVI} = [H_x, H_y, \hat{V}]$ is concatenated with the 3D geometric priors ($\mathbf{P}$ and $\mathbf{N}$) to construct the initial multi-modal embedding $\mathbf{F}_{0} \in \mathbb{R}^{C \times H \times W}$:
\begin{equation}
    \mathbf{F}_{0} = \mathcal{E}_{fuse} \Big( \hat{\mathbf{I}}_{HVI} \oplus \mathcal{E}_{geo}(\mathbf{P} \oplus \mathbf{N}) \Big)
\end{equation}
where $\oplus$ denotes channel-wise concatenation, and $\mathcal{E}_{geo}, \mathcal{E}_{fuse}$ are shallow convolutional projection blocks. This unified initialization ensures that $\mathbf{F}_{0}$ possesses both noise-robust chromaticity and explicit 3D spatial awareness.

\subsection{Physics-Guided Feature Extraction}
\label{subsec:physics_guided_extraction}

Implicit self-attention often struggles to distinguish intrinsic dark textures from actual shadows. To overcome this, we propose the \textbf{Geometric \& Semantic Dual Explicit Guided Attention (GSGA)} mechanism. This module explicitly modulates the attention affinity matrix using 3D planar correlations and high-level semantic correlations, injecting physical laws directly into the feature aggregation process.

Given the intermediate feature map $\mathbf{F}_{in} \in \mathbb{R}^{C \times H \times W}$, we first project it into Query ($\mathbf{Q}$), Key ($\mathbf{K}$), and Value ($\mathbf{V}$) tensors. The standard self-attention computes the affinity matrix via the dot product $\mathbf{Q}\mathbf{K}^T$. In our GS-DEGA, we introduce two explicit guidance matrices: the Geometric Guidance Matrix $\mathbf{G}_{geo}$ and the Semantic Guidance Matrix $\mathbf{G}_{sem}$.

\noindent \textbf{Geometric Guidance via 3D Surface Normals.} 
According to Lambertian reflectance theory, regions with similar surface orientations under the same light source should exhibit highly correlated illumination degradation. To ensure computational feasibility and avoid the $O(H^2W^2)$ memory bottleneck of global attention, GSGA operates within non-overlapping local windows of size $M \times M$ (we set $M=8$). Consequently, the guidance matrices are strictly constrained to a highly efficient $\mathbb{R}^{M^2 \times M^2}$. For any two spatial locations $i$ and $j$ within the same local window, the planar correlation is computed using the cosine similarity of their corresponding 3D surface normals $\mathbf{n}_i, \mathbf{n}_j \in \mathbf{N}$. The Geometric Guidance Matrix $\mathbf{G}_{geo} \in \mathbb{R}^{M^2 \times M^2}$ is defined as:
\begin{equation}
    \mathbf{G}_{geo}(i, j) = \max \left( 0, \frac{\mathbf{n}_i \cdot \mathbf{n}_j}{\|\mathbf{n}_i\| \|\mathbf{n}_j\|} \right)^{\gamma}
\end{equation}
where $\gamma$ is a hyperparameter controlling the strictness of the geometric constraint. This matrix explicitly encourages the network to aggregate features from co-planar regions within the local neighborhood, preventing the erroneous blending of features across sharp 3D geometric boundaries (e.g., occlusion edges).

\noindent \textbf{Semantic Guidance via DINO Priors.} 
While geometric guidance ensures physical consistency, it may over-smooth distinct textures that happen to lie on the same plane. To preserve texture fidelity, we compute the Semantic Guidance Matrix $\mathbf{G}_{sem} \in \mathbb{R}^{M^2 \times M^2}$ within the identical local window using the frozen DINOv2 features $\mathbf{F}_{dino}$ extracted in Sec.~\ref{subsec:prior_estimation}:
\begin{equation}
    \mathbf{G}_{sem}(i, j) = \text{Softmax}_j \left( \frac{\mathbf{f}_{dino, i} \cdot \mathbf{f}_{dino, j}}{\sqrt{C_{dino}}} \right)
\end{equation}
where $\mathbf{f}_{dino, i}$ is the semantic token at location $i$. DINO features are inherently robust to illumination variations, thus $\mathbf{G}_{sem}$ provides an illumination-invariant semantic affinity.

\noindent \textbf{Dual Explicit Modulation.} 
The final Dual Explicit Guided Attention is formulated by element-wise multiplying ($\odot$) the standard affinity matrix with our dual guidance matrices before applying the Value aggregation:
\begin{equation}
    \text{GSGA}(\mathbf{Q}, \mathbf{K}, \mathbf{V}) = \text{Softmax} \left( \frac{\mathbf{Q}\mathbf{K}^T}{\sqrt{d_k}} \odot \mathbf{G}_{geo} \odot \mathbf{G}_{sem} \right) \mathbf{V}
\end{equation}
This formulation mathematically guarantees that strong attention weights are assigned only to regions that are both geometrically co-planar and semantically consistent, effectively bridging the 2D-3D domain gap.

\noindent \textbf{Global Semantic Injection via CLIP.} 
Despite robust localized feature extraction, deeply shadowed regions often suffer from irreversible information degradation. To address this ill-posed hallucination problem, we introduce a global semantic injection module at the bottleneck of the U-shaped backbone. Specifically, we utilize a frozen CLIP visual encoder to extract a highly abstract, holistic semantic token $\mathbf{t}_{clip} \in \mathbb{R}^{1 \times C}$ from the input image. 

To propagate this global prior into the local feature space, we employ a Cross-Modal Attention mechanism. Let $\mathbf{F}_{bottle} \in \mathbb{R}^{N \times C}$ denote the flattened bottleneck feature map (where $N = H \times W$). $\mathbf{F}_{bottle}$ serves as the query ($\mathbf{Q}$), while the CLIP token $\mathbf{t}_{clip}$ acts as both the key ($\mathbf{K}$) and value ($\mathbf{V}$). The semantically enriched bottleneck feature $\mathbf{F}_{bottle}^{\prime}$ is formulated as:
\begin{equation}
    \mathbf{F}_{bottle}^{\prime} = \mathbf{F}_{bottle} + \text{Softmax}\left(\frac{(\mathbf{F}_{bottle}\mathbf{W}_q)(\mathbf{t}_{clip}\mathbf{W}_k)^\top}{\sqrt{d_k}}\right) (\mathbf{t}_{clip}\mathbf{W}_v)
\end{equation}
where $\mathbf{W}_q, \mathbf{W}_k, \mathbf{W}_v \in \mathbb{R}^{C \times d_k}$ are learnable linear projection matrices, and $d_k$ is the scaling factor. This explicitly guides the network to hallucinate plausible textures based on holistic scene comprehension.

\subsection{Frequency Collaborative Reconstruction Module (FCRM)}
\label{subsec:fcrm}

Monolithic decoders struggle to balance sharp occlusion boundaries (high-frequency) and smooth illumination variations (low-frequency). The FCRM overcomes this by explicitly disentangling the deep feature $\mathbf{F}_{d}$ into $\mathbf{F}_{high}$ and $\mathbf{F}_{low}$ branches.

\noindent \textbf{Frequency-Aware Decoupling Strategy.} 
Let $\mathbf{F}_{d}$ denote the deep feature representation obtained from the encoder. The FCRM first explicitly splits $\mathbf{F}_{d}$ into high-frequency ($\mathbf{F}_{high}$) and low-frequency ($\mathbf{F}_{low}$) branches, allowing for specialized architectural designs tailored to each sub-band's characteristics.

\noindent \textbf{High-Frequency Texture Branch.} 
To capture localized, sharp structural details and occlusion boundaries, we employ a Dense CNN block for $\mathbf{F}_{high}$. Convolutional operations, with their strong inductive bias for local spatial translation invariance, are highly effective at reconstructing high-frequency gradients. By progressively concatenating intermediate features to ensure maximum information flow, the high-frequency residual $\mathbf{R}_{high}$ is formulated as:
\begin{equation}
    \mathbf{R}_{high} = \mathbf{W}_{proj} \ast \Big[ \mathbf{F}_{high}, \mathcal{F}_1(\mathbf{F}_{high}), \dots, \mathcal{F}_L \big( [ \mathbf{F}_{high}, \dots, \mathbf{F}_{high}^{(L-1)} ] \big) \Big]
\end{equation}
where $\mathcal{F}_l(\cdot)$ denotes the convolution and non-linear activation at the $l$-th layer, $[\cdot]$ represents channel-wise concatenation, and $\ast$ is the convolution operator.

\noindent \textbf{Low-Frequency Illumination Branch.} 
Global illumination consistency requires long-range dependency modeling. We process $\mathbf{F}_{low}$ using a Mini Transformer U-Net integrated with Haar Wavelet transforms. The Discrete Wavelet Transform (DWT) acts as a mathematically rigorous downsampler preserving low-frequency energy. Specifically, the DWT operation $\mathcal{W}(\cdot)$ decomposes $\mathbf{F}_{low} \in \mathbb{R}^{C \times H \times W}$ into four sub-bands (LL, LH, HL, HH), yielding a compact tensor of shape $\mathbb{R}^{4C \times \frac{H}{2} \times \frac{W}{2}}$. To focus on global illumination, we process this representation using a Mini Transformer U-Net $\mathcal{T}_{unet}$, which efficiently models long-range lighting contexts with reduced computational overhead. Finally, the inverse DWT ($\mathcal{W}^{-1}(\cdot)$) restores the spatial resolution to $\mathbb{R}^{C \times H \times W}$. The low-frequency illumination map is mathematically reconstructed as:
\begin{equation}
    \mathbf{R}_{low} = \mathcal{W}^{-1} \Big( \mathcal{T}_{unet} \big( \mathcal{W}(\mathbf{F}_{low}) \big) \Big)
\end{equation}
This formulation ensures that the illumination manifold is explicitly smoothed and refined within the specialized wavelet sub-bands.

\noindent \textbf{Geometry-Conditioned Collaborative Fusion.} 
Finally, the decoupled representations are collaboratively fused. Crucially, we continuously inject the 3D surface normal prior $\mathbf{N}$ during fusion to ensure the restored gradients strictly adhere to planar geometries:
\begin{equation}
    \mathbf{I}_{out} = \mathcal{M}_{fuse} \big( \mathbf{R}_{high} \oplus \mathbf{R}_{low}, \mathbf{N} \big) + \mathbf{I}_{in}
\end{equation}
where $\oplus$ denotes feature concatenation, $\mathcal{M}_{fuse}$ is a projection mapping block, and the global residual connection $+\mathbf{I}_{in}$ stabilizes the network by focusing optimization entirely on learning the degradation residual. This frequency-collaborative architecture ensures a rigorous balance between photorealistic detail hallucination and physical illumination consistency. 

\begin{figure*}[t]
    \centering
    \includegraphics[width=\textwidth]{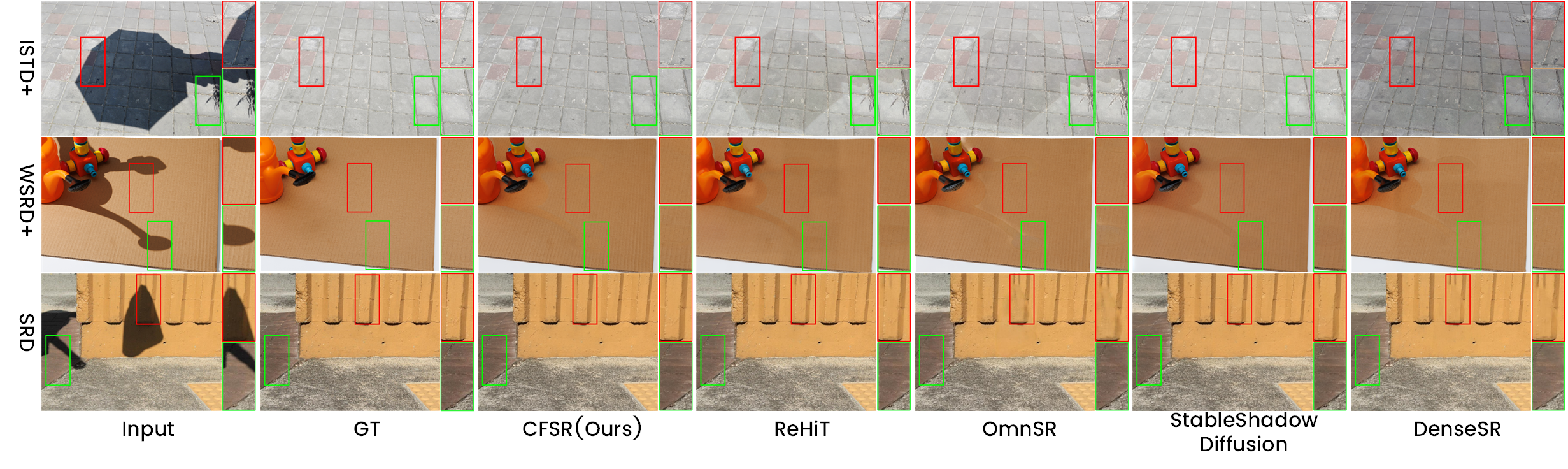}
    \captionsetup{skip=-2pt}
    \caption{
    \textbf{Qualitative comparisons on the ISTD+, WSRD+, and SRD datasets.} Our CFSR effectively removes shadows while strictly preserving background textures and color consistency, successfully avoiding the boundary artifacts and residual shadows frequently produced by other state-of-the-art methods.
}
    \label{fig:compare_others}
\end{figure*}

\section{Experiment}
\begin{table*}[!t]
\small
\center
\scalebox{0.9}{
\begin{tabular}{l c c c c c c c c c c c c c}
\hline
\multirow{2}{*}{Method} & \multirow{2}{*}{Venue} & \multicolumn{2}{c}{ISTD} & \multicolumn{2}{c}{ISTD+} & \multicolumn{2}{c}{SRD} & \multicolumn{2}{c}{INS} & \multicolumn{2}{c}{WSRD+} & \multicolumn{2}{c}{Ambient6K} \\
\cmidrule(lr){3-4}\cmidrule(lr){5-6}\cmidrule(lr){7-8}\cmidrule(lr){9-10}\cmidrule(lr){11-12}\cmidrule(lr){13-14}
& & {PSNR$\uparrow$} & {SSIM$\uparrow$} & {PSNR$\uparrow$} & {SSIM$\uparrow$} & {PSNR$\uparrow$} & {SSIM$\uparrow$} & {PSNR$\uparrow$} & {SSIM$\uparrow$} & {PSNR$\uparrow$} & {SSIM$\uparrow$} & {PSNR$\uparrow$} & {SSIM$\uparrow$} \\
\hline
DHAN~\cite{cun2020towards} & AAAI 2020 & 29.11 & 0.954 & 25.66 & 0.956 & 30.02 & 0.899 & 27.84 & 0.963 & 22.39 & 0.796 & --- & --- \\

DC-ShadowNet~\cite{jin2021dc} & ICCV 2021 & 24.02 & 0.677 & 25.50 & 0.694 & 30.91 & 0.915 & 26.19 & 0.921 & 21.62 & 0.593 & 17.73 & 0.711 \\
BMNet~\cite{zhu2022bijective} & CVPR 2022 & 28.53 & 0.952 & 32.22 & 0.965 & 28.34 & 0.943 & 27.90 & 0.958 & 24.75 & 0.816 & --- & --- \\
ShadowFormer~\cite{Dong_2024_CVPR} & AAAI 2023 & 29.90 & 0.960 & 31.39 & 0.946 & 30.58 & 0.958 & 28.62 & 0.963 & 25.44 & 0.820 & --- & --- \\
DMTN~\cite{liu2023decoupled} & TMM 2023 & 29.05 & 0.956 & 31.72 & 0.963 &  32.45 & 0.964 & 28.83 & 0.969 & --- & --- & --- & --- \\
ShadowDiffusion~\cite{guo2023shadowdiffusion} & CVPR 2023 & 30.09 & 0.918 & 31.08 & 0.950 & 31.91 & 0.968 & 29.12 & 0.966 & --- & --- & --- & --- \\
Refusion~\cite{Refusion} & CVPRW 2023 & 29.34 & 0.951 & 32.41 & 0.961 & 31.60 & 0.949 & 28.13 & 0.958 & 22.32 & 0.738 & --- & --- \\
ShadowRefiner~\cite{dong2024shadowrefiner} & CVPRW 2024 & 28.75 & 0.916 & 31.03 & 0.928 & --- & --- & --- & --- & 26.04 & 0.827 & --- & --- \\
IFBlend~\cite{vasluianu2024towards} & ECCV 2024 & 28.55 & 0.906 & 30.87 & 0.916 & 31.95 & 0.950 & --- & --- & 25.79 & 0.809 & 21.44 & 0.819 \\
DeS3 & AAAI 2024 & 29.15 & 0.947 & 31.39 & 0.957  & \textbf{34.11} & 0.968  & 27.89 & 0.947 & --- & --- & --- & ---\\ 
RLN$^2$-Lf~\cite{vasluianu2025after} & ICCV 2025 & 28.77 & 0.914 & 31.02 & 0.930 & --- & --- & --- & --- & 25.84 & 0.821 & 21.71 & 0.825 \\
ReHiT~\cite{dong2025retinex} & CVPRW 2025 & 28.81 & 0.914 & 31.16 & 0.925 & --- & --- & --- & --- & 26.15 & 0.826 & 19.98 & 0.798 \\
OmniSR~\cite{xu2025omnisr} & AAAI 2025 & 30.45 & 0.964 & 33.34 & \underline{0.970} & 32.87 & \underline{0.969} & 30.38 & 0.973 & 26.07 & 0.835 & 23.01 & 0.830 \\
StableShadowDiffusion \cite{xu2025detail} & CVPR 2025 & --- & --- & \textbf{35.19} & \underline{0.970} & \underline{33.63} & 0.968 & 30.56 & \underline{0.975} & 26.26 & 0.827 & --- & --- \\
DenseSR~\cite{lin2025densesr} & ACMMM 2025 & 30.64 & \underline{0.976} & 33.98 & \textbf{0.974} & 33.45 & \textbf{0.970} & \underline{30.64} & \textbf{0.981} & 26.28 & 0.838 & 22.54 & 0.826 \\
PhaSR~\cite{phasr}  & CVPR 2026 & \underline{30.73} & 0.960 & 34.48 & 0.960 & --- & --- & 30.38 & 0.961 & \textbf{28.44} & \textbf{0.942} & \underline{23.32} & \underline{0.834} \\
\rowcolor[HTML]{E7E6E6}
CFSR(Ours) & --- & \textbf{30.91} & \textbf{0.979} & \underline{34.93} & \textbf{0.974} & \underline{33.63} & \underline{0.969} & \textbf{30.72} & \underline{0.975} & \underline{27.37} & \underline{0.901} & \textbf{23.62} & \textbf{0.843} \\
\hline
\end{tabular}}
\caption{\textbf{Quantitative comparisons on shadow removal and ambient lighting normalization benchmarks.} We benchmark the proposed CFSR against state-of-the-art methods across five standard datasets (ISTD~\cite{wang2018stacked}, ISTD+~\cite{le2019shadow}, SRD~\cite{SRD}, INS~\cite{xu2025omnisr}, WSRD+~\cite{vasluianu2023wsrd}) and the challenging Ambient6K~\cite{vasluianu2024towards}, which features complex multi-illumination and indirect lighting. Our CFSR consistently achieves superior performance, demonstrating its robust generalization across diverse ambient lighting scenarios. \textbf{Bold} and \underline{underlined} indicate the best and second-best results.}
\vspace{-0.2cm}
\label{tab:comparison_ISTD}
\end{table*}

\subsection{Experimental Settings}

\noindent \textbf{Datasets and Evaluation.} 
We comprehensively evaluate the proposed CFSR on several widely adopted shadow removal benchmarks, including the ISTD~\cite{wang2018stacked}, ISTD+~\cite{le2019shadow}, SRD\cite{SRD}, INS~\cite{xu2025omnisr}, WSRD+~\cite{vasluianu2023wsrd}, and Ambient6K~\cite{vasluianu2024towards} datasets. Following previous methods~\cite{fu2021auto,guo2023shadowformer,le2020shadow}, we evaluate the images at a resolution of $256 \times 256$ via random cropping. To quantify the restoration performance, we adopt standard evaluation protocols in the shadow removal literature and employ two widely used metrics: Peak Signal-to-Noise Ratio (PSNR) and the Structure Similarity Index Measure (SSIM). For fair comparison, the quantitative results of competing methods are either directly cited from their original papers or evaluated using their official implementations with default hyperparameters.

\noindent \textbf{Training Details.} 
The proposed framework is implemented in PyTorch and trained from scratch using $8 \times$ NVIDIA RTX 4090 GPUs. During training, we randomly crop $360 \times 360$ patches from the original images and apply standard data augmentations, including random horizontal and vertical flips. The network is optimized using the AdamW optimizer ($\beta_1 = 0.9, \beta_2 = 0.999$, weight decay $= 0.02$) to minimize the Charbonnier loss. The total training process spans $1400$ epochs. We employ a Cosine Annealing learning rate scheduler, where the initial learning rate is set to $2 \times 10^{-4}$ with a $3$-epoch linear warmup, gradually decaying to a minimum of $5 \times 10^{-5}$. The batch size is set to $8$ per GPU, resulting in an effective total batch size of $64$. To facilitate future research, our code and models will be released upon acceptance.

\noindent \textbf{Quantitative Results.} 
Table~\ref{tab:comparison_ISTD} summarizes the quantitative comparisons between our CFSR and $15$ recent state-of-the-art methods across six diverse benchmarks. CFSR consistently delivers top-tier performance, securing the best or second-best results across almost all metrics. On the widely used ISTD dataset, our method achieves a state-of-the-art PSNR of 30.91~dB and an SSIM of 0.979. Most notably, on the highly challenging Ambient6K benchmark, which features complex multi-source indirect illumination, CFSR significantly outperforms existing approaches, attaining a PSNR of 23.62~dB and an SSIM of 0.843. Although concurrent models such as StableShadowDiffusion and DenseSR show marginal advantages on specific datasets like ISTD+ or SRD, CFSR maintains highly competitive performance across the board. This validates the robust generalization capability of our framework in handling both traditional hard shadows and complex ambient lighting scenarios without dataset-specific tuning.

\begin{table}[t]
\centering
\small
\setlength{\tabcolsep}{3.5pt}
\renewcommand{\arraystretch}{1.05}
\begin{tabular}{l|cc|cc|cc}
\toprule
\multirow{2}{*}{Method} 
& \multicolumn{2}{c|}{ISTD+ $\rightarrow$ SRD} 
& \multicolumn{2}{c|}{SRD $\rightarrow$ ISTD+} 
& \multicolumn{2}{c}{INS $\rightarrow$ WSRD+} \\
& PSNR & SSIM & PSNR & SSIM & PSNR & SSIM \\
\midrule
ShadowDiffusion & 24.26 & 0.890 & 29.48 & 0.947 & 19.40 & 0.825 \\
Refusion        & 22.56 & 0.886 & 27.31 & 0.924 & 18.73 & 0.787 \\
IFBlend         & 26.23 & 0.923 & 30.23 & 0.953 & 20.11 & 0.828 \\
OmniSR          & 25.86 & 0.917 & 30.19 & 0.951 & 20.03 & 0.825 \\
\rowcolor[HTML]{E7E6E6}
CFSR(Ours)      & \textbf{27.02} & \textbf{0.931} & \textbf{30.75} & \textbf{0.958} & \textbf{20.36} & \textbf{0.848} \\
\bottomrule
\end{tabular}
\caption{\textbf{Cross-dataset evaluation.} ISTD+ $\rightarrow$ SRD means training on the ISTD+ dataset and testing on the SRD dataset.}
\label{tab:cross_dataset}
\vspace{-0.5cm}
\end{table}

\noindent \textbf{Qualitative Evaluation.} 
Figure~\ref{fig:compare_others} provides visual comparisons against leading methods on the ISTD+, WSRD+, and SRD datasets. While existing methods achieve reasonable quantitative metrics, they frequently introduce perceptual flaws in complex regions. For example, ReHiT and OmniSR struggle with subtle penumbra transitions, often leaving distinct boundary artifacts or residual shadow traces. Generative approaches like StableShadowDiffusion can occasionally cause color degradation or over-smooth fine textural details within the restored areas. Conversely, CFSR accurately harmonizes the illumination between shadow and non-shadow regions. As highlighted in the close-up patches, our method excels at eradicating boundary artifacts while faithfully preserving the underlying structural textures and original chromaticity, yielding outputs that are visually closest to the ground truth.

\noindent \textbf{Cross-Dataset Generalization.} 
To verify that CFSR learns physically meaningful priors rather than overfitting to dataset-specific biases, we conduct cross-dataset evaluations (Table~\ref{tab:cross_dataset}). Our model is trained on a source dataset and directly evaluated on an unseen target dataset without fine-tuning. Remarkably, CFSR demonstrates superior generalization across all settings. For instance, in the challenging INS $\rightarrow$ WSRD+ transfer, CFSR surpasses the state-of-the-art OmniSR by 0.33~dB in PSNR. This robust transferability confirms that our explicit 3D geometric constraints and global semantics provide illumination-invariant structural guidance, effectively mitigating domain shifts.

\subsection{Analysis and Discussion}

\begin{table}[t]
\centering
\resizebox{0.9\linewidth}{!}{
\begin{tabular}{l | cc}
\toprule
\textbf{Configuration of Priors} & \textbf{PSNR} $\uparrow$ & \textbf{SSIM} $\uparrow$ \\
\midrule
w/o Priors (Base) & 20.90 & 0.836 \\
\midrule
Semantic Prior (CLIP only) & 22.45 & 0.863 \\
Semantic Prior (DINO only) & 27.21 & 0.898 \\
Geometry Prior (Point+Normal only) & 21.37 & 0.840 \\
\midrule
\rowcolor[HTML]{E7E6E6}
\textbf{CLIP + DINO + Geometry (Ours)} & 
\textbf{27.37} & \textbf{0.901} \\
\bottomrule
\end{tabular}
}
\caption{Ablation study on multi-modal priors. The results demonstrate that semantic priors (CLIP, DINO) and geometric guidance (Point+Normal) provide orthogonal benefits, and their joint integration is crucial for optimal performance.}
\label{tab:ablation_priors_overall}
\vspace{-0.2cm}
\end{table}

\begin{figure}[t!]
\vspace{-0.2cm}
\centering
\captionsetup{skip=-2pt}
\includegraphics[width=1.0\linewidth]{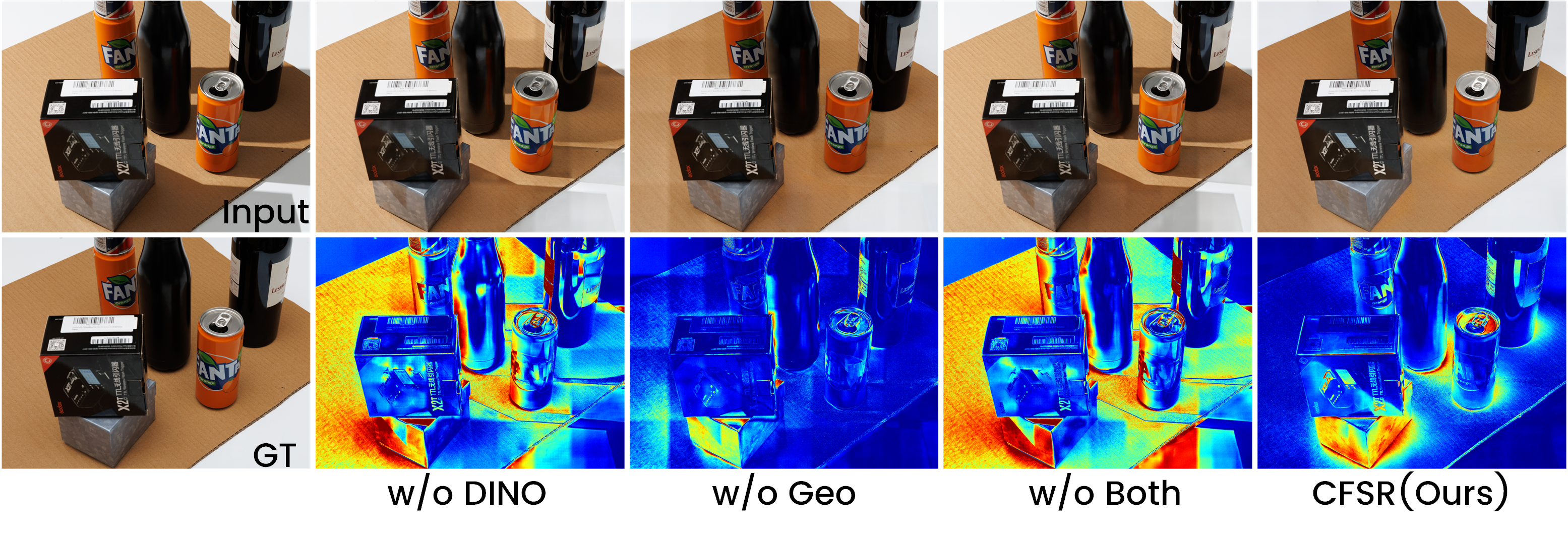}
\caption{\textbf{Visual ablation of multi-modal priors.} }
\label{fig:abl_priors} 
\vspace{-0.2cm}
\end{figure}

\noindent \textbf{Ablation Study on Multi-modal Priors.} 
We ablate the proposed multi-modal priors on the WSRD+ dataset, as reported in Table~\ref{tab:ablation_priors_overall}. Removing all priors causes a severe performance collapse. The dense semantic prior (DINO) contributes most significantly to the quantitative metrics by providing robust region-level context. Building upon this, the global prior (CLIP) and spatial geometry prior yield further steady gains, culminating in the best overall performance. This quantitative trend is strongly corroborated by visual evidence (Figure~\ref{fig:abl_priors}). The semantic priors ensure global color fidelity; without DINO, severe color shifts occur. Conversely, the geometry prior is essential for preserving local 3D structural consistency. Without it (\textit{w/o Geo}), residual shadows and irregular illumination boundaries persist on complex curved surfaces, such as the cans and base blocks.

\noindent \textbf{Decoupling Analysis.}
To further understand their specific contributions, we evaluate the priors on two decoupled subsets: \textbf{Class A} (flat regions with complex textures) and \textbf{Class B} (shadows cast on curved surfaces/folds). As shown in Table~\ref{tab:ablation_decoupling}, the semantic prior dominates Class A, where its removal drops PSNR by 2.34 dB compared to the full model. Interestingly, removing the geometry prior slightly improves Class A. This suggests that estimated depth/normal cues may introduce minor interference when recovering pure 2D textures. 
However, in Class B scenarios, the geometry prior is crucial. It improves PSNR by 0.25 dB by explicitly guiding the illumination deformation over 3D curved surfaces, a task where purely semantic features struggle. This trade-off analysis demonstrates that our design effectively balances the texture continuity from semantics and structural adaptation from geometry.

\begin{table}[t]
\centering
\resizebox{\linewidth}{!}{
\begin{tabular}{l | cc | cc}
\toprule
\multirow{2}{*}{\textbf{Variant}} & \multicolumn{2}{c|}{\textbf{Class A)}} & \multicolumn{2}{c}{\textbf{Class B}} \\
\cmidrule(lr){2-3} \cmidrule(lr){4-5}
& \textbf{PSNR} $\uparrow$ & \textbf{SSIM} $\uparrow$ & \textbf{PSNR} $\uparrow$ & \textbf{SSIM} $\uparrow$ \\
\midrule
w/o Both (Baseline) & 17.86 & 0.687 & 21.53 & 0.916 \\
w/o Semantic (DINO) & 18.16 & 0.708 & 21.90 & 0.916 \\
w/o Geometry        & \textbf{20.74} & 0.759 & 29.60 & 0.967 \\
Full Model          & 20.50 & \textbf{0.760} & \textbf{29.85} & \textbf{0.968} \\
\bottomrule
\end{tabular}
}
\caption{Decoupling analysis of multi-modal priors. The semantic prior dominates texture-rich flat regions (Class A), while the geometry prior is critical for illumination recovery on 3D curved surfaces (Class B).}
\label{tab:ablation_decoupling}
\vspace{-0.2cm}
\end{table}

\begin{figure}[t!]
\centering
\captionsetup{skip=-2pt}
\includegraphics[width=1.0\linewidth]{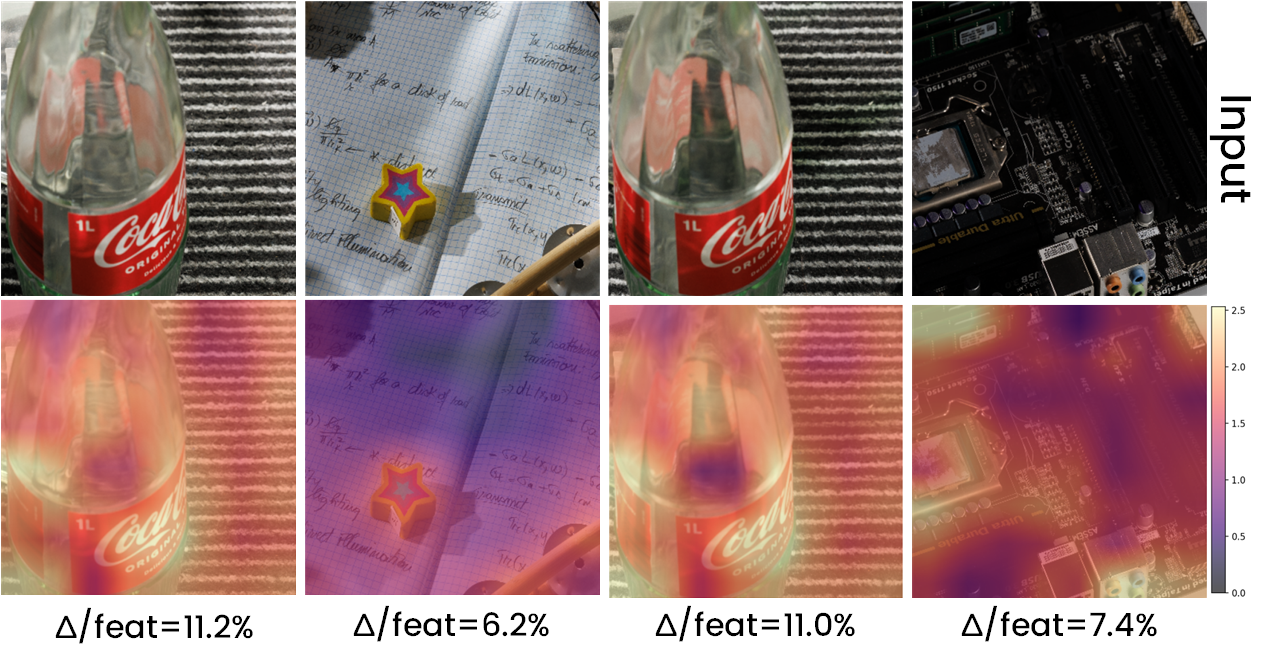} 
\caption{\textbf{Feature impact of CLIP semantic injection on Ambient6K dataset~\cite{vasluianu2024towards}.} The heatmaps show the relative feature modulation magnitude ($\Delta/\text{feat}$) induced by the CLIP prior at the bottleneck. Higher intensity (yellow) indicates stronger activation. }
\label{fig:eff_clip}
\vspace{-0.2cm}
\end{figure}

\noindent \textbf{Effectiveness of Global Semantic Injection.} 
While DINO provides dense patch-level semantics and geometry provides structural constraints, heavily shadowed regions often suffer from irreversible local information degradation. To combat this, CFSR injects holistic context via a frozen CLIP encoder at the bottleneck. To demystify its contribution, we visualize the feature impact heatmap in Figure~\ref{fig:eff_clip}. The colormap visualizes the relative feature modulation magnitude ($\Delta\text{feature}_2$) induced by the Cross-Modal Attention mechanism. As explicitly observed, the CLIP prior exhibits remarkably strong activations (highlighted in yellow/orange) within deeply occluded areas (e.g., the dark penumbra behind the star block or inside the curved bottle) and complex structural intersections (e.g., the motherboard circuitry). This phenomenon robustly validates our design intuition: when local receptive fields encounter near-zero pixel values, the network dynamically relies on CLIP's global abstract semantics to plausibly hallucinate and recover the lost textures.

\begin{figure}[t!]
\centering
\vspace{-0.2cm}
\captionsetup{skip=-2pt}
\includegraphics[width=1.0\linewidth]{figs/freq_vis.png}
\caption{\textbf{Decoupled frequency feature comparison.} Low-frequency (left) and high-frequency (right) features from DenseSR~\cite{lin2025densesr} and our CFSR. Our FCRM produces a cleaner decomposition with minimal shadow leakage.}
\label{fig:freq_vis}
\vspace{-0.2cm}
\end{figure}

\noindent \textbf{Effectiveness of FCRM.} 
Figure~\ref{fig:freq_vis} compares the frequency representations of our FCRM with DenseSR~\cite{lin2025densesr}. DenseSR exhibits shadow residuals in low-frequency maps and entangled artifacts in high-frequency maps. In contrast, FCRM yields a much cleaner decomposition. The low-frequency features suppress shadow responses via content-aware adaptive filtering, while high-frequency features concentrate tightly along object boundaries without shadow contamination. This neat decoupling preserves sharp boundaries while preventing the color bleeding typical of competing methods.

We quantitatively validate this in Table~\ref{tab:fcrm_ablation}. Replacing FCRM with vanilla concatenation yields 26.58 dB. Adding either the low-frequency ($\mathcal{L}$) or high-frequency ($\mathcal{H}$) branch individually improves global illumination or boundary fidelity, respectively. Operating both branches jointly with our frequency-aware fusion (FCRMFusionCore) achieves the best result, demonstrating the necessity of adaptive frequency routing for high-quality restoration.

\begin{table}[t]
\centering
\resizebox{0.95\linewidth}{!}{
\begin{tabular}{ccc|cc|l}
\toprule
$\mathcal{L}$ & $\mathcal{H}$ & Fusion & \textbf{PSNR} $\uparrow$ & \textbf{SSIM} $\uparrow$ & \textbf{Role} \\
\midrule
\xmark & \xmark & \xmark & 26.58 & 0.887 & Vanilla skip connection \\
\cmark & \xmark & \xmark & 26.89 & 0.891 & Global illumination \\
\xmark & \cmark & \xmark & 26.74 & 0.895 & Boundary sharpening \\
\cmark & \cmark & \xmark & 27.12 & 0.897 & Complementary gains \\
\cmark & \cmark & \cmark & \textbf{27.37} & \textbf{0.901} & Full FCRM \\
\bottomrule
\end{tabular}
}
\caption{\textbf{Component ablation within FCRM.} $\mathcal{L}$: low-freq branch; $\mathcal{H}$: high-freq branch; Fusion: frequency-aware skip fusion.}
\label{tab:fcrm_ablation}
\vspace{-0.4cm}
\end{table}

\section{Conclusion}
In this paper, we propose CFSR, a novel multi-modal prior-driven framework that reframes shadow removal from an unconstrained 2D image-to-image translation task into a physics-conditioned restoration process. By explicitly leveraging 3D geometric cues (depth and normals) and robust semantics (DINO and CLIP), CFSR effectively mitigates severe information degradation in shadowed regions. Specifically, we introduce the Geometric \& Semantic Dual Explicit Guided Attention (GSGA) to structurally constrain feature aggregation using physical and semantic correlations. Furthermore, our Frequency Collaborative Reconstruction Module (FCRM) disentangles representations to achieve an optimal balance between high-fidelity texture recovery and global illumination smoothing. Extensive experiments demonstrate that CFSR achieves state-of-the-art performance across multiple challenging benchmarks, yielding physically consistent and boundary-sharp shadow-free images. We believe our paradigm of integrating foundation model priors with explicit physical constraints offers valuable insights for broader low-level vision tasks.

\clearpage
\bibliographystyle{ACM-Reference-Format}
\bibliography{main}


\end{document}